\documentclass[10pt,twocolumn,letterpaper]{article}

\usepackage[pagenumbers]{cvpr} 
\usepackage[dvipsnames]{xcolor}

\definecolor{cvprblue}{rgb}{0.21,0.49,0.74}
\usepackage[pagebackref,breaklinks,colorlinks,citecolor=cvprblue]{hyperref}

\def\paperID{*****} 
\def\confName{CVPR}
\def\confYear{2024}

\newlength\savewidth\newcommand\shline{\noalign{\global\savewidth\arrayrulewidth
  \global\arrayrulewidth 1pt}\hline\noalign{\global\arrayrulewidth\savewidth}}
\title{\emph{Diffusion-RWKV}: \\ Scaling RWKV-Like Architectures for Diffusion Models}

\author{Zhengcong Fei, Mingyuan Fan, Changqian Yu\\
Debang Li, Junshi Huang*\\
Kunlun Inc.\\
{\tt\small feizhengcong@gmail.com}
}

\begin{document}

\twocolumn[{%
\renewcommand\twocolumn[1][]{#1}%
\maketitle

\vspace{-1.cm}
\begin{center}
    \centering
    \captionsetup{type=figure}
    \includegraphics[width=1\textwidth]{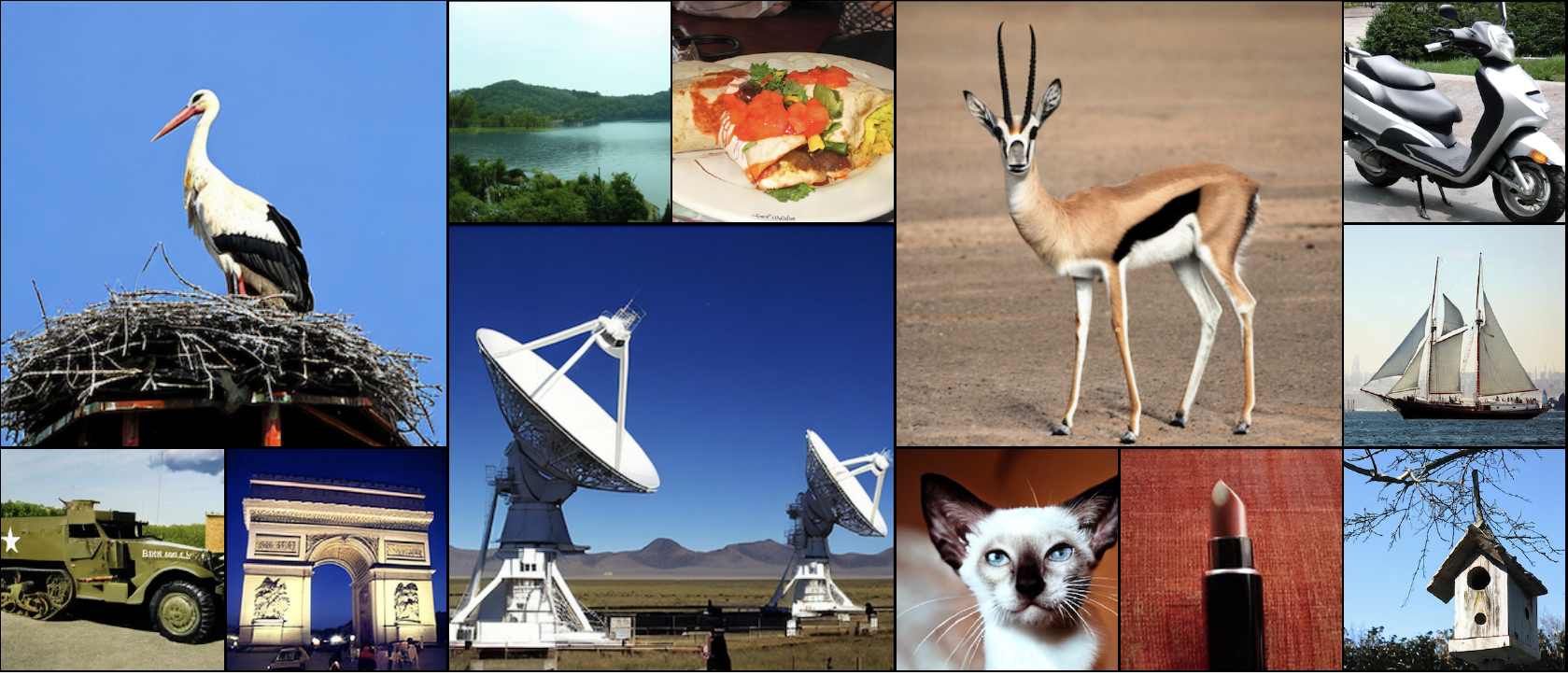}
    \captionof{figure}{
    \textbf{Diffusion models with RWKV-like backbones achieve comparable image quality.} 
     Selected samples generated by class-conditional Diffusion-RWKV trained on the ImageNet with resolutions of 256$\times$256 and 512$\times$512, respectivly.
     %
    }
    \label{fig:vim_teaser}
\end{center}%
}]

\begin{abstract}
Transformers have catalyzed advancements in computer vision and natural language processing (NLP) fields. However, substantial computational complexity poses limitations for their application in long-context tasks, such as high-resolution image generation. This paper introduces a series of architectures adapted from the RWKV model used in the NLP, with requisite modifications tailored for diffusion model applied to image generation tasks, referred to as Diffusion-RWKV. Similar to the diffusion with Transformers, our model is designed to efficiently handle patchnified inputs in a sequence with extra conditions, while also scaling up effectively, accommodating both large-scale parameters and extensive datasets. Its distinctive advantage manifests in its reduced spatial aggregation complexity, rendering it exceptionally adept at processing high-resolution images, thereby eliminating the necessity for windowing or group cached operations. Experimental results on both condition and unconditional image generation tasks demonstrate that Diffison-RWKV achieves performance on par with or surpasses existing CNN or Transformer-based diffusion models in FID and IS metrics while significantly reducing total computation FLOP usage.
\end{abstract}    
\section{Introduction}

Transformers \cite{vaswani2017attention,radford2019language,mann2020language,dosovitskiy2020image,he2022masked,lin2022survey}, which have gained prominence due to their adaptable nature and proficient information processing capabilities, have set new standards across various domains including computer vision and NLP. Notably, they have demonstrated exceptional performance in tasks like image generation \cite{parmar2018image,ding2021cogview,ding2022cogview2,lee2022autoregressive,ramesh2021zero,chang2022maskgit,fei2022progressive,peebles2023scalable}. However, the self-attention operation in Transformer exhibits a quadratic computational complexity, thereby limiting their efficiency in handling long sequences and poses a significant obstacle to their widespread application \cite{kitaev2020reformer,tay2022efficient,zhou2021informer,zamir2022restormer,fei2022progressive}. Consequently, there is a pressing need to explore architectures that can effectively harness their versatility and robust processing capabilities while mitigating the computational demands. It becomes even more crucial in the context of high-resolution image synthesis or the generation of lengthy videos.

\begin{figure*}[t]
  \centering
   \includegraphics[width=0.9\linewidth]{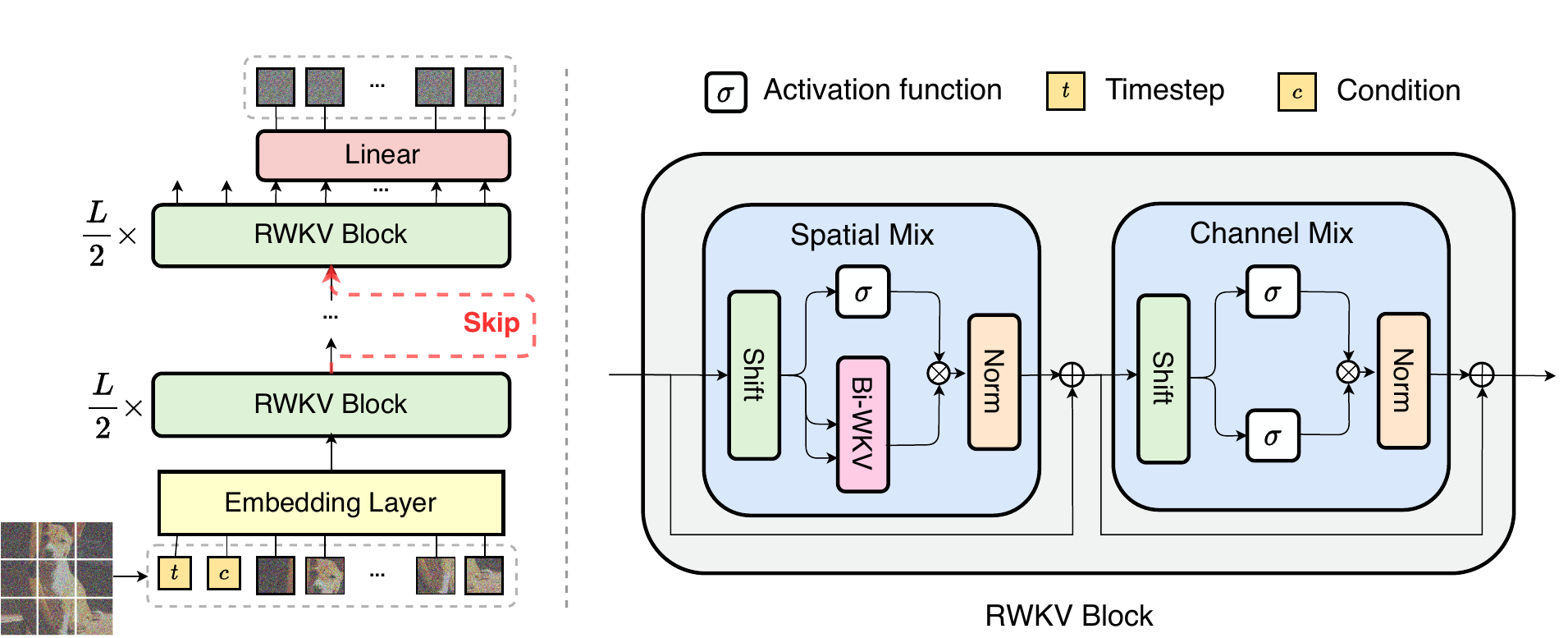}
   \caption{\textbf{Overall framework of diffusion models with RWKV-like architectures.}  (a) The Diffusion-RWKV architecture comprises $L$ identical Bi-RWKV layers, a patch embedding, and a projection layer. A skip connection is established between shallow and deep stacked Bi-RWKV layers for information flow. (b) The detailed composition of Bi-RWKV layers, includes a shift method and a bidirectional RNN cell in spatial mix, and a shift with two activate functions in channel mix. }
   \label{fig:framework} 
\end{figure*}

In recent developments, models such as RWKV \cite{peng2023rwkv} and Mamba \cite{gu2023mamba}, have emerged as popular solutions for enhancing efficiency and processing lengthy textual data with comparable capacity. These innovative models exhibit characteristics akin to transformers
\cite{brown2020language,devlin2018bert,lewis2019bart,liu2019roberta,radford2018improving,radford2019language,smith2022using,stickland2019bert,fei2022deecap,yan2021semi,fei2021partially}, encompassing to handle long-range dependencies and parallel processing. Moreover, they have demonstrated scalability, performing admirably with large-scale NLP and CV datasets \cite{zhu2024vision,fei2024scalable}. However, given the substantial dissimilarities between visual and textual data domains, it remains challenging to envision complete replacement of Transformers with RWKV-based methods for vision generation tasks \cite{duan2024vision}. 
It becomes imperative to conduct an in-depth analysis of how these models are applied to image generation tasks. This analysis should investigate their scalability in terms of training data and model parameters, evaluate their efficiency in handling visual data sequentially, and identify the essential techniques to ensure model stability during scaling up.

This paper introduces Diffusion-RWKV, which is designed to adapt the RWKV architecture in diffusion models for image generation tasks. The proposed adaptation aims to retain the fundamental structure and advantages of RWKV \cite{peng2023rwkv} while incorporating crucial modifications to tailor it specifically for synthesizing visual data. 
Specifically, we employ Bi-RWKV \cite{duan2024vision} for backbone, which enables the calculation within linear computational complexity in an RNN form forward and backward. 
We primarily make the architectural choices in diffusion models, including condition incorporation, skip connection, and finally offer empirical baselines that enhance the model’s capability while ensuring scalability and stability. 
Building on the aforementioned design,  a diverse set of Diffusion-RWKV models is developed, as a broad range of model scales, ranging from tiny to large. 
These models are training on CIFAR-10, Celebrity to ImageNet-1K using unconditional and class-conditioned training at different image resolutions. Moreover, performance evaluations are conducted in both raw and latent spaces. Encouragingly, under the same settings, Diffusion-RWKV has comparable performance to competitor DiT \cite{peebles2023scalable} in image generation, with lower computational costs while maintaining stable scalability. 
This achievement enables Diff-RWKV training parallelism, high flexibility, excellent performance, and low inference cost simultaneously, making it a promising alternative in image synthesis. 
The contribution can be summarized as: 
\begin{itemize}
\item In a pioneering endeavor, we delve into the exploration of a purely RWKV-based diffusion model for image generation tasks, positioning as a low-cost alternative to Transformer. 
Our model not only inherits the advantages of RWKV for long-range dependency capture, but also reduces complexity to a linear level. 
\item To cater to the demands of image synthesis, we have conducted a comprehensive and systematic investigation of Diffusion-RWKV models by exploring various configuration choices pertaining to conditioning, block design, and model parameter scaling. 
\item Experimental results indicate that Diffusion-RWKV performs comparably to well-established benchmarks DiTs and U-ViTs, exhibiting lower FLOPs and faster processing speeds as resolution increases. Notably, Diffusion-RWKV achieves a 2.95 FID score trained only on ImageNet-1k. Code and model are available at \url{https://github.com/feizc/Diffusion-RWKV}. 
\end{itemize}

\section{Methodology}

This section commences by providing an overview of the foundational concepts in Section \ref{sec21}. Subsequently, we delve into a comprehensive exposition of the RWKV-based diffusion models for image generation in Section \ref{sec22}. It encompasses various aspects such as image patchnify, stacked Bi-RWKV block, skip connections, and condition incorporation. Lastly, we perform computational analysis and establish optimal model scaling configurations.

\subsection{Preliminaries}
\label{sec21}

\paragraph{Diffusion models.}

Diffusion models have emerged as a new family of generative models that generate data by iterative transforming random noise through a sequence of deconstructible denoising steps. It usually includes a forward noising process and a backward denoising process. Formally, given data $x_0$ sampled from the distribution $p(x_0)$, the forward noising process involves iteratively adding Gaussian noise to the data, creating a Markov Chain of latent variables $x_1, \ldots, x_T$, where:
\begin{equation}
q(x_t|x_{t-1}) = \mathcal{N}(x_t; \sqrt{1 - \beta_t} x_{t-1}, \beta_t {I}),
\end{equation}
and $\beta_1, \ldots, \beta_T$ are hyperparameters defining the noise schedule. 
After a pre-set number of diffusion steps, $x_T$ can be considered as standard Gaussian noise. 
A denoising network $\epsilon_\theta$ with parameters $\theta$ is trained to learn the backward denoising process, which aims to remove the added noise according to a noisy input. 
During inference, a data point can be generated by sampling from a random Gaussian noise $x_T \sim \mathcal{N}(0; {I})$ and iteratively denoising the sample by sequentially sampling $x_{t-1}$ from $x_t$ with the learned denoising process, as:
\begin{equation}
x_{t-1} = \frac{1}{\sqrt{\alpha_t}}(x_t - \frac{1 - \alpha_t}{1 - \overline{\alpha}_t} \epsilon(x_t, t)) + \sigma_t z,
\end{equation}
where $\overline{\alpha}t = \prod{s=1}^t \alpha_s$, $\alpha_t = 1 - \beta_t$, and $\sigma_t$ denotes the noise scale. In practice, the diffusion sampling process can be further accelerated with various sampling techniques \cite{lu2022dpm,song2020denoising,lu2022dpm+}.

\paragraph{RWKV-like structures.}

RWKV \cite{peng2023rwkv} brought improvements for standard RNN architecture \cite{hochreiter1997long}, which is computed in parallel during training while inference like RNN. 
It involves enhancing the linear attention mechanism and designing the receptance weight key value (RWKV) mechanism. 
Generally, RWKV model consists of an input layer, a series of stacked residual blocks, and an output layer. Each residual block is composed of time-mix and channel-mix sub-block. 

\textbf{(i)} The Time-Mix Block aims to improve the modeling of dependencies and patterns within a sequence. It is achieved by replacing the conventional weighted sum calculation in an attention mechanism with hidden states. The time-mix block can effectively propagate and updates information across sequential steps with hidden states and the updation can be expressed as follows:
\begin{align}
    q_t &=  ({\mu}_{q} \odot x_{t} + (1 -{\mu}_{q} ) \odot x_{t-1}) \cdot  W_q , \\
    k_t &=  ({\mu}_{k} \odot x_{t} + (1 -{\mu}_{k} ) \odot x_{t-1}) \cdot  W_k , \\
    v_t &=  ({\mu}_{v} \odot x_{t} + (1 -{\mu}_{v} ) \odot x_{t-1}) \cdot  W_v , \\
    o_t &=  (\sigma (q_t) \odot h(k_t, v_t)) \cdot  W_o   ,
\end{align}
where $q_t$, $k_t$, and $v_t$ are calculated by linearly interpolating between the current input and the input at the previous time step. 
The interpolation, determined by the token shift parameter $\mu$, ensures coherent and fluent token representations. 
Additionally, a non-linear activation function $\sigma$ is applied to $q_t$, and the resulting value is combined with the hidden states $h(k_t, v_t)$ using element-wise multiplication. The hidden states, which serve as both the reset gate and a replacement for the traditional weighted sum value, can be computed as: 
\begin{align}
    p_t &= \max(p_{t-1}, k_t), \\
    h_t &= \frac{\text{exp}(p_{t-1}-p_t) \odot a_{t-1} + \text{exp}(k_t - p_t) \odot v_t}{\text{exp}(p_{t-1}-p_t) \odot b_{t-1} + \text{exp}(k_t - p_t)} ,
\end{align}
where $a_0, b_0, p_0$ are zero-initialized. Intuitively, the hidden states are computed recursively, and the vector $p$ serves as the reset gate in this process. 

\textbf{(ii)} Channel-Mix Block aims to amplify the outputs of time-mix block, which can be given by:
\begin{align}
    r_t &= ({\mu}_r \odot o_t + (1-{\mu}_r) \odot o_{t-1}) \cdot  W_r \\
    z_t &=  ({\mu}_z \odot o_t + (1-{\mu}_z) \odot o_{t-1}) \cdot  W_z \\
    \tilde{x}_t &= \sigma (r_t) \odot (\max(z_t, 0)^2 \cdot  W_v)
\end{align}
The output $o_t$ contains historical information up to time $t$, and the interpolation weight $\mu$ is derived from $o_t$ and $o_{t-1}$, similar to the time-mix block, which also enhances the historical information representation.
Note that the calculations of hidden states may lead to information loss and failure to capture long-range dependencies \cite{peng2023rwkv}.

\begin{table}[t]
\centering
\scalebox{1.}{
\begin{tabular}{l|ccccc}
& \#Params & $L$  &$D$ & $E$ &Gflops  \\ \shline
Small  & 38.9M &25 & 384 &4& 1.72 \\
Base &74.3M&25 &768&4& 3.32 \\
Medium &132.0M &49 & 768 &4 &5.90  \\
Large &438.5M&49&1024&4&19.65 \\
Huge &779.1M&49 &1536&4& 34.95
\\ 
\end{tabular}}
\caption{\textbf{Scaling law model size.} The model sizes and detailed hyperparameter settings for scaling experiments. In between, $L$ is the number of stacked Bi-RWKV layers, $D$ is the hidden state size, and $E$ is the embedding ratio. 
}
\label{tab:scale}
\end{table}

\subsection{Model Structure Design} 
\label{sec22}
We present Diffusion-RWKV, a variant of RWKV-like diffusion models, as a simple and versatile architecture for image generation.
Diffusion-RWKV parameterizes the noise prediction network $\epsilon_\theta({x}_t, t, {c})$, which takes the timestep $t$, condition ${c}$ and noised image ${x}_t$ as inputs and predicts the noise injected into data point ${x}_t$. 
As our goal follows the cutting-edge RWKV architecture to maintain its scalability characteristics, Diffusion-RWKV is grounded in the bidirectional RWKV \cite{duan2024vision} architecture which operates on sequences of tokens. Figure \ref{fig:framework} illustrates an overview of the complete Diffusion-RWKV architecture. In the following, we elaborate on the forward pass and the components that constitute the design space of this model class.

\paragraph{Image tokenization.}
The initial layer of Diffusion-RWKV performs a transformation of the input image ${I} \in \mathbb{R}^{H \times W \times C}$ into flattened 2-D patches ${X} \in \mathbb{R}^{J \times (p^2 \cdot C)}$. Subsequently, it converts these patches into a sequence of $J$ tokens, each with $D$ dimension, by linearly embedding each image patch in the input. Consistent with \cite{dosovitskiy2020image}, learnable positional embeddings are applied to all input tokens. The number of tokens $J$ generated by the tokenization process is determined by the hyperparameter patch size $p$, calculated as $\frac{H \times W}{p^2}$. The tokenization layer supports both raw pixel and latent space representations.

\paragraph{Bi-directional RWKV block.}
Subsequent to the embedding layer, the input tokens undergo processing through a succession of identical Bi-RWKV blocks. 
Considering that the original RWKV block was designed for one-dimensional sequence processing, we resort to \cite{zhu2024vision}, which incorporates bidirectional sequence modeling tailored for vision tasks. This adaptation preserves the core structure and advantages of RWKV \cite{peng2023rwkv} while integrating critical modifications to tailor it for processing visual data. Specifically, it employs a quad-directional shift operation tailored for two-dimensional vision data and modifies the original causal RWKV attention mechanism to a bidirectional global attention mechanism. The quad-directional shift operation expands the semantic range of individual tokens, while the bidirectional attention enables the calculation of global attention within linear computational complexity in an RNN-like forward and backward manner. As illustrated in the right part of Figure \ref{fig:framework}, the forward pass of Bi-RWKV blocks amalgamates both forward and backward directions in the spatial and channel mix modules. These alterations enhance the model's long-range capability while ensuring scalability and stability.

\paragraph{Skip connection.}
Considering a series of $L$ stacked Bi-RWKV blocks, we categorize the blocks into three groups: the first $\lfloor \frac{L}{2} \rfloor$ blocks as the shallow group, the middle block as the central layer, and the remaining $\lfloor \frac{L}{2} \rfloor$ blocks as the deep group as \cite{fei2024scalable,bao2023all}. Let $h_{shallow}$ and $h_{deep}$ represent the hidden states from the main branch and long skip branch, respectively, both residing in $\mathbb{R}^{J \times D}$. We propose concatenating these hidden states and applying a linear projection, expressed as \texttt{Linear(Concate(}$h_{shallow}, h_{deep}$\texttt{))}, before propagating them to the subsequent block.

\paragraph{Linear decoder.}
Upon completion of the final Bi-RWKV block, it becomes essential to decipher the sequence of hidden states to generate an output noise prediction and diagonal covariance prediction. These resulting outputs maintain a shape equivalent to the original input image. To achieve this, a standard linear decoder is employed, wherein the final layer norm is applied, and each token is linearly decoded into a $p^2 \cdot C$ tensor. Finally, the decoded tokens are rearranged to match their original spatial layout, yielding the predicted noise and covariance.

\begin{figure*}[t]
  \centering
   \includegraphics[width=1.\linewidth]{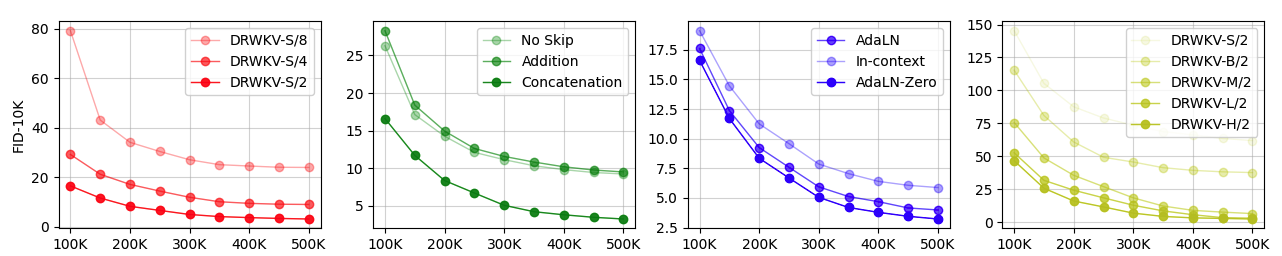}
   \caption{\textbf{Ablation experiments and model analysis for different designs with DRWKV-S/2 model on the CIFAR10 dataset.} We report FID metrics on 10K generated samples every 50K steps. We can find that: (a) {Patch size}. A smaller patch size can improve the image generation performance. (b) {Skip operation}. Combining the long skip branch can accelerate the training as well as optimize generated results.  (c) {Variants of condition incorporation}. AdaLN-Zero is an effective strategy for conditioning. (d) {Model parameters scaling}. As we expected, holding the patch size constant, increasing the model parameters can consistently improve the generation performance. 
   }
   \label{fig:ab} 
\end{figure*}

\paragraph{Condition incorporation.}
In addition to the noised image inputs, diffusion models process supplementary conditional information, such as noise timesteps $t$ and condition $\textbf{c}$, which usually encompass class labels or natural language data.
In order to incorporate additional conditions effectively, this study employs three distinct designs as referred from \cite{peebles2023scalable,fei2024scalable}:
\begin{itemize}
    \item \emph{In-context conditioning.} A straightforward strategy of appending the vector embeddings of timestep $t$ and condition $\textbf{c}$ as two supplementary tokens within the input sequence. These tokens are treated on par with the image tokens. Implementing this technique allows for the utilization of Bi-RWKV blocks without requiring any adjustments.  Note that the conditioning tokens are removed from the sequence in the spatial mix module in each Bi-RWKV block and after the final block.
    \item \emph{Adaptive layer norm (adaLN) block.} We explore replacing the standard norm layer with adaptive norm layer. Rather than directly learning scale and shift parameters on a per-dimension basis, these parameters are deduced from the summation of the embedding vectors of $t$ and $\textbf{c}$.
    \item \emph{adaLN-Zero block.} In addition to regressing $\gamma$ and $\beta$, we also regress dimension-wise scaling parameters $\alpha$ that are applied immediately prior to any residual connections within the Bi-RWKV block. The MLP is initialized to produce a zero-vector output for all $\alpha$. 
\end{itemize}

\subsection{Computation Analysis}

In summary, the hyper-parameters of the Diffusion-RWKV model encompass crucial components including embedding dimension $E$, hidden dimension $D$ in linear projection, and depth $L$. 
Central to the bi-directional RWKV block's architecture is the generation of attention results for each token through individual update steps, culminating in the requisite 
$T$ steps for the complete processing of the WKV matrix. 
Here, $T$ is the sequence length.
Considering the input $K$ and $V$ are matrices with the shape of $J \times D$, where $D$ is the dimension of hidden learnable vectors, the computational cost of calculating the WKV matrix is given by:
\begin{equation}
     \label {equation:attn_flops} \text {FLOPs}(\text {Bi\mbox {-}WKV}(K, V)) = 13 \times J \times D. 
\end{equation}
Here, the number 13 is approximately from the updates of four hidden states, the computation of the exponential, and the calculation of wkvt matrix. $J$ is the total number of update steps and is equal to the number of image tokens. The above approximation shows that the complexity of the forward process is $O(J \cdot D)$. The backward propagation of the operator can still be represented as a more complex RNN form, with a computational complexity of $O(J \cdot D)$. It demonstrates a superiority of linear increasing compared with self-attention operation in Transformer structure.
Finally, the different model variants are specified in Table \ref{tab:scale}. In between, we use five configs, from small to huge, to cover a wide range of model sizes and flop allocations, from 1.72 to 34.95 Gflops, allowing us to gauge scaling performance.

\section{Experiments}

In this section, we shall delve into the intricacies of the design space and thoroughly examine the scaling properties inherent in our Diffusion-RWKV model class. In order to simplify our discourse, each model within our class is denoted by its specific configurations and patch size $p$. As an example, we refer to the Large version configuration with $p=2$ as DRWKV-L/2.

\subsection{Experimental Settings}

\paragraph{Datasets.}
For unconditional image generation, two datasets are considered: CIFAR10 \cite{krizhevsky2009learning} and CelebA 64x64 \cite{liu2015deep}. CIFAR10 comprises a collection of 50k training images, while CelebA 64x64 encompasses 162,770 images depicting human faces. As for the class-conditional image generation, the ImageNet dataset \cite{deng2009imagenet} is employed. This dataset consists of 1,281,167 training images, distributed across 1,000 distinct classes. In terms of data augmentation, only horizontal flips are employed.
The training process involves 500k iterations on both CIFAR10 and CelebA 64$\times$64, utilizing a batch size of 128 in pixel space. In the case of ImageNet, two scenarios are considered as resolution of 256$\times$256 and 512$\times$512. For the former, 500k iterations are conducted, while for the latter, 1M iterations are performed. The batch size is set as 512 in both cases.

\paragraph{Implementation details. }
We followed the same training recipe from DiT \cite{peebles2023scalable} to ensure consistent settings across all models. We choose to incorporate an exponential moving average (EMA) of model weights with a fixed decay rate of 0.9999. All reported results have been obtained using the EMA model. 
We use the AdamW optimizer \cite{kingma2014adam} without weight decay across all datasets and maintain a learning rate of 1e-4 to 3e-5 in stages. Our models are trained on the Nvidia A100 GPU. 
During training on the ImageNet dataset at a resolution of 256$\times$256 and 512$\times$512, we also adopt classifier-free guidance \cite{ho2022classifier} following \cite{rombach2022high} and use an off-the-shelf pre-trained variational autoencoder (VAE) model \cite{kingma2013auto} from playground V2 provided in huggingface\footnote{https://huggingface.co/playgroundai} with corresponding settings.
The VAE encoder component incorporates a downsampling factor of 8.
We maintain the diffusion hyperparameters from \cite{peebles2023scalable}, employing a $t_{max}=1000$ linear variance schedule ranging from $1\times10^{-4}$ to $2\times10^{-2}$ and parameterization of the covariance. 
In order to adapt the model to an unconditional context, we just removed the class label embedding component.

\paragraph{Evaluation metrics.}
The performance evaluation of image generation is conducted using the Fréchet Inception Distance (FID) \cite{heusel2017gans}, an extensively employed metric for assessing the quality of generated images. In accordance with established conventions for comparative analysis with previous works, we present FID-50K results obtained through 250 DDPM sampling steps \cite{parmar2022aliased}, following \cite{dhariwal2021diffusion}. Furthermore, we provide supplementary metrics such as the Inception Score \cite{salimans2016improved}, sFID \cite{nash2021generating}, and Precision/Recall \cite{kynkaanniemi2019improved} to complement the evaluation.

\subsection{Model Analysis}
We first conduct a systematical empirical investigation into the fundamental components of Diffusion-RWKV models. Specifically, we ablate on the CIFAR10 dataset, evaluate the FID score every 50K training iterations on 10K generated samples, instead of 50K samples for efficiency \cite{bao2023all}, and determine the optimal default implementation details.

\paragraph{Effect of patch size.}

We train patch size range over (8, 4, 2) in Small configuration on the CIFAR10 dataset. 
The results obtained from this experimentation are illustrated in Figure \ref{fig:ab} (a). It indicates that the FID metric exhibits fluctuations in response to a decrease in patch size while maintaining a consistent model size.
Throughout the training process, we observed discernible improvements in FID values by augmenting the number of tokens processed by Diffusion-RWKV, while keeping the parameters approximately fixed. This observation leads us to the conclusion that achieving optimal performance requires a smaller patch size, specifically 2.
We hypothesize that this requirement arises from the inherently low-level nature of the noise prediction task in diffusion models. It appears that smaller patches are more suitable for this task compared to higher-level tasks such as classification.

\paragraph{Effect of long skip.}

To assess the efficacy of the skipping operation, we investigate three different variants, namely: ($\textbf{i}$) Concatenation, denoted as \texttt{Linear(Concat(}$h_{shallow}$, $h_{deep}$\texttt{))}; ($\textbf{ii}$) Addition, represented by $h_{shallow} + h_{deep}$; and ($\textbf{iii}$) No skip connection.
Figure \ref{fig:ab} (b) illustrates the outcomes of these variants. It is evident that directly adding the hidden states from the shallow and deep layers does not yield any discernible benefits. Conversely, the adoption of concatenation entails a learnable linear projection on the shallow hidden states and effectively enhances performance in comparison to the absence of a long skip connection.

\begin{table}[t]
\small
	\begin{center}
	\setlength{\tabcolsep}{2mm}{
		\begin{tabular}{lcc}
		Model    &\#Params &FID$\downarrow$   \\ \shline
DDPM \cite{ho2020denoising} &36M& 3.17 \\
EDM \cite{karras2022elucidating} &56M& 1.97 \\ 
GenViT \cite{yang2022your} &11M& 20.20  \\
U-ViT-S/2 \cite{bao2023all} & 44M& 3.11 \\ 
DiS-S/2 \cite{fei2024scalable} &28M &3.25 \\ \hline 
DRWKV-S/2 &39M & 3.03 \\
		\end{tabular}
  {\caption{\textbf{Benchmarking unconditional image generation on CIFAR10}. Diffusion-RWKV-S/2 model obtains comparable results with fewer parameters. }
			\label{tab:2}}
			}
	\end{center}
\end{table}

\begin{table}[t]
\small
	\begin{center}
	\setlength{\tabcolsep}{2mm}{
		\begin{tabular}{lcc}
		Model &\#Params  &FID$\downarrow$ \\ \shline
DDIM \cite{song2020denoising} &79M & 3.26    \\
Soft Trunc. \cite{kim2021soft} &62M & 1.90 \\
U-ViT-S/4 \cite{bao2023all}& 44M & 2.87\\
DiS-S/2 \cite{fei2024scalable} & 28M &2.05 \\ \hline
DRWKV-S/2 &39M & 1.92 \\
		\end{tabular}
  {\caption{\textbf{Benchmarking unconditional image generation on CelebA 64$\times$64}. Diffusion-RWKV-S/2 maintains a superior generation performance in small model settings.}
			\label{tab:3}}
			}
	\end{center}
\end{table}

\paragraph{Effect of condition combination.}
We examine three approaches for incorporating the conditional timestep $t$ into the network, as discussed in the preceding method section. The integration methods are depicted in Figure \ref{fig:ab} (c). Among these strategies, the adaLN-Zero block exhibits a lower FID compared to the in-context conditioning approach, while also demonstrating superior computational efficiency. Specifically, after 500k training iterations, the adaLN-Zero model achieves an FID that is approximately one-third of that obtained by the in-context model, underscoring the critical influence of the conditioning mechanism on the overall quality of the model. Furthermore, it should be noted that the initialization process holds significance in this context.
Additionally, it is worth mentioning that due to the inclusion of a resize operation in the design of the Bi-RWKV in spatial channel mix, only the in-context token is provided to the channel mix module.

\paragraph{Scaling model size.}

We investigate scaling properties of Diffusion-RWKV by studying the effect of depth, \emph{i.e.}, number of Bi-RWKV layers, and width, e.g. the hidden size. Specifically, we train 5 Diffusion-RWKV models on the ImageNet dataset with a resolution of 256$\times$256, spanning model configurations from small to huge as detailed in Table \ref{tab:scale}, denoted as (S, B, M, L, H) for simple. 
As depicted in Figure \ref{fig:ab} (d), the performance improves as the depth increases from 25 to 49. Similarly, increasing the width from 384 to 1024 yields performance gains. 
Overall, across all five configurations, we find that similar to DiT models \cite{peebles2023scalable}, large models use FLOPs
more efficient and scaling the DRWKV will improve the FID at all stages of training.

\subsection{Main Results}

\begin{table}[t]
\centering %
    \small
    \scalebox{0.85}{
    \begin{tabular}{lccccc}
    Model & FID$\downarrow$   & sFID$\downarrow$  & IS$\uparrow$     & Precision$\uparrow$ & Recall$\uparrow$ \\
      \shline
    BigGAN-deep~\cite{brock2018large} & 6.95 & 7.36 & 171.4 & 0.87 & 0.28 \\
    StyleGAN-XL~\cite{sauer2022stylegan} & 2.30 & 4.02 & 265.12 & 0.78 & 0.53 \\
    ADM~\cite{dhariwal2021diffusion} & 10.94 & 6.02 & 100.98 & 0.69 & 0.63 \\
    ADM-U & 7.49 & 5.13 & 127.49 & 0.72 & 0.63 \\
    ADM-G & 4.59 & 5.25 & 186.70 & 0.82 & 0.52 \\
    ADM-G, ADM-U & 3.94 & 6.14      & 215.84 & 0.83 & 0.53 \\
    CDM~\cite{ho2022cascaded}  & 4.88 & - & 158.71 & - & - \\
    LDM-8~\cite{rombach2022high} & 15.51 & - & 79.03 & 0.65 & 0.63 \\
    LDM-8-G & 7.76 & - & 209.52 & 0.84 & 0.35 \\
    LDM-4 & 10.56 & - & 103.49 & 0.71 & 0.62 \\
    LDM-4-G & 3.60 & - & 247.67 &  {0.87} & 0.48 \\
    VDM++ \cite{kingma2023understanding} &2.12 &-&267.70 & - & -\\
     U-ViT-H/2 \cite{bao2023all} &2.29 &5.68&263.88 &0.82 & 0.57 \\
     DiT-XL/2 \cite{peebles2023scalable}   &  {2.27} &  {4.60} &  {278.24} & 0.83 & 0.57 \\
     SiT-XL/2 \cite{ma2024sit} & 2.06 & 4.50 & 270.27&0.82 &0.59\\
     DiffuSSM-XL/2 \cite{yan2023diffusion} &2.28 & 4.60& 278.24 &0.83 &0.57 \\
     DiS-H/2 \cite{fei2024scalable}  &2.10 &4.55 & 271.32 &0.82 & 0.58 \\
    \hline 
    DRWKV-H/2 & 2.16 &4.58 & 275.36 &0.83 &0.58 \\
    \end{tabular}}
    \caption{\textbf{Benchmarking class-conditional image generation on ImageNet 256$\times$256.} Diffusion-RWKV-H/2 achieves state-of-the-art FID metrics towards best competitors.}
    \label{tab:4}
\end{table}

We compare to a set of previous best models, includes: GAN-style approaches that previously achieved state-of-the-art results, UNet-architectures trained with pixel space representations, and Transformers and state space models operating in the latent space.
Note that our aim is to compare, through a similar denoising process, the performance of
our model with respect to other baselines.

\paragraph{Unconditional image generation.}

We evaluate the unconditional image generation capability of our model in relation to established baselines using the CIFAR10 and CelebA datasets within the pixel-based domain. The outcomes of our analysis are presented in Table \ref{tab:2} and Table \ref{tab:3}, respectively. The results reveal that our proposed model, Diffusion-RWKV, attains FID scores comparable to those achieved by Transformer-based U-ViT and SSM-based DiS models, while utilizing a similar training budget. Notably, our model achieves this with fewer parameters and exhibits superior FID scores. These findings emphasize the practicality and effectiveness of RWKV across various image generation benchmarks.

\paragraph{Class-conditional image generation.}

We also compare the Diffusion-RWKV model with state-of-the-art class-conditional models in the ImageNet dataset, as listed in Table \ref{tab:4} and Table \ref{tab:5}. When considering a resolution of 256, the training of our DRWKV model exhibits a 25\% reduction in Total Gflops compared to the DiT (1.60$\times$10$^{11}$ vs. 2.13$\times$10$^{11}$). Additionally, our models achieve similar sFID scores to other DDPM-based models, outperforming most state-of-the-art strategies except for SiT and DiS. This demonstrates that the images generated by the Diffusion-RWKV model are resilient to spatial distortion. Furthermore, in terms of FID score, Diffusion-RWKV maintains a relatively small gap compared to the best competitor.
It is noteworthy that SiT is a transformer-based architecture that employs an advanced strategy, which could also be incorporated into our backbone. However, this aspect is left for future research, as our primary focus lies in comparing our model against DiT.
Moreover, we extend our comparison to a higher-resolution benchmark of size 512. The results obtained from the Diffusion-RWKV model demonstrate a relatively strong performance, approaching that of some state-of-the-art high-resolution models. Our model outperforms all models except for DiS, while achieving comparable FID scores with a lower computational burden.

\subsection{Case Study}
\begin{figure*}[t]
  \centering
   \includegraphics[width=0.96\linewidth]{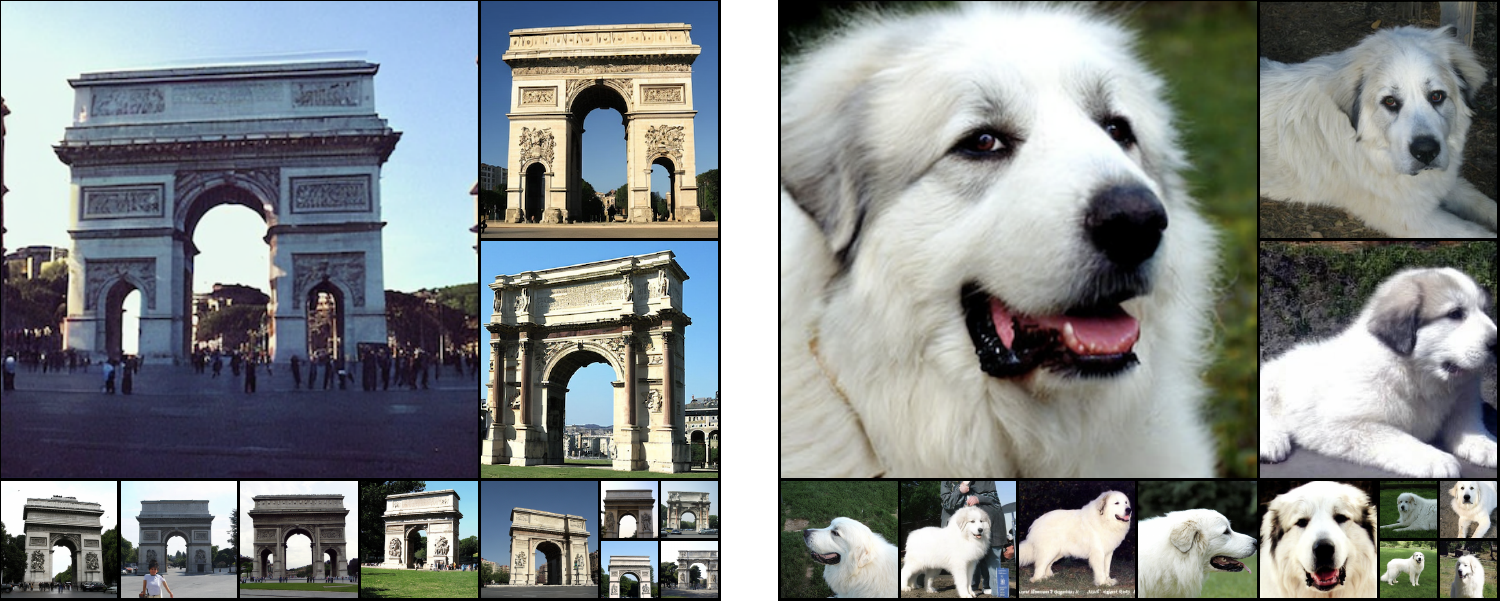}
   \caption{\textbf{Image results generated from Diffusion-RWKV model. } Selected samples on ImageNet 512$\times$512 with sample classes and different seeds. We can see that Diffusion-RWKV can generate high-quality images while keeping integrated condition alignment. 
   }
   \label{fig:cases} 
\end{figure*}

In Figure \ref{fig:vim_teaser} and Figure \ref{fig:cases}, a curated selection of samples from the ImageNet datasets is presented. These samples are showcased at resolutions of 256$\times$256 and 512$\times$512, effectively illustrating clear semantic representations and exhibiting high-quality generation. To delve deeper into this topic, the project page offers a collection of additional generated samples, encompassing both class-conditional and random variations.

\begin{table}[t]
\centering
\small
    \scalebox{0.85}{
    \begin{tabular}{lccccc}
    Model & FID$\downarrow$   & sFID$\downarrow$  & IS$\uparrow$     & Precision$\uparrow$ & Recall$\uparrow$ \\
     \shline
    BigGAN-deep~\cite{brock2018large} & 8.43 & 8.13 & 177.90 & 0.88 & 0.29 \\
    StyleGAN-XL~\cite{sauer2022stylegan} & 2.41 & 4.06 & 267.75 & 0.77 & 0.52 \\
    ADM~\cite{dhariwal2021diffusion} & 23.24 & 10.19 & 58.06 & 0.73 & 0.60  \\
    ADM-U & 9.96 & 5.62 & 121.78 & 0.75 &   {0.64} \\
    ADM-G & 7.72 & 6.57 & 172.71 &   {0.87} & 0.42 \\
    ADM-G, ADM-U & 3.85 & 5.86 & 221.72 & 0.84 & 0.53\\
    VDM++ \cite{kingma2023understanding} &2.65 &-&278.10 & - & -\\
     U-ViT-H/4 \cite{bao2023all} &4.05 &6.44 &263.79&0.84 &0.48 \\
      {DiT-XL/2} \cite{peebles2023scalable} &   {3.04} &   {5.02} &   {240.82} & 0.84 & 0.54 \\
      DiffuSSM-XL/2 \cite{yan2023diffusion} &3.41 & 5.84 & 255.06 & 0.85 & 0.49\\
     DiS-H/2 \cite{fei2024scalable} & 2.88 &4.74  &272.33 & 0.84 &0.56 \\ \hline
     DRWKV-H/2  & 2.95 & 4.95& 265.20 & 0.84 & 0.54\\ 
    \end{tabular}}
    \caption{\textbf{Benchmarking class-conditional image generation on ImageNet 512$\times$512.} DRWKV-H/2 demonstrates a promising performance compared with both CNN-based and Transformer-based UNet for diffusion.} %
    \label{tab:5}
\end{table}

\section{Related Works}

\paragraph{Image generation with diffusion.}
Diffusion and score-based generative models \cite{hyvarinen2005estimation,song2019generative,song2020score,song2023consistency} have demonstrated significant advancements in various tasks, particularly in the context of image generation \cite{ramesh2022hierarchical,rombach2022high,saharia2022photorealistic}. The DDPM has been primarily attributed to improvements in sampling techniques \cite{ho2020denoising,karras2022elucidating,nichol2021improved,fei2023gradient,fei2023uncertainty}, and the incorporation of classifier-free
guidance \cite{ho2022classifier}. Additionally, \cite{song2020denoising} introduced a more efficient sampling procedure called Denoising Diffusion Implicit Model (DDIM). Latent space modeling is another core technique in deep generative models. Variational autoencoders \cite{kingma2013auto} pioneered learning latent spaces with encoder-decoder architectures for reconstruction. 
The concept of compressing information in latent spaces was also adopted in diffusion models, as exemplified by the state-of-the-art sample quality achieved by latent diffusion models \cite{rombach2022high}, which train deep generative models to reverse a noise corruption process within a latent space. Additionally, recent advancements have incorporated masked training procedures, enhancing denoising training objectives through masked token reconstruction \cite{zheng2023fast}.
Our work is fundamentally built upon existing standard DDPMs.

\paragraph{Architectures for diffusion models.}
Early models for diffusion employed U-Net style architectures \cite{dhariwal2021diffusion,ho2020denoising}. Subsequent studies endeavored to enhance U-Nets by incorporating various techniques, such as the addition of attention layers at multiple scales \cite{nichol2021improved}, residual connections \cite{brock2018large}, and normalization \cite{perez2018film,wu2018group}. However, U-Nets encounter difficulties when scaling to high resolutions due to the escalating computational demands imposed by the attention mechanism \cite{shaham2018spectralnet}. Recently, vision transformers \cite{dosovitskiy2020image} have emerged as an alternative architecture, showcasing their robust scalability and long-range modeling capabilities, thereby challenging the notion that convolutional inductive bias is always indispensable. 
Diffusion transformers \cite{bao2023all,peebles2023scalable,fei2019fast} demonstrated promising results. Other hybrid CNN-transformer architectures were proposed \cite{liu2021swin} to improve training stability.
More recently, state space-based model \cite{hu2024zigma,fei2024scalable,yan2023diffusion} have obtain a advanced performance with computation efficiency. 
Our work aligns with the exploration of recurrent sequence models and the associated design choices for generating high-quality images while mitigating text similarity.

\paragraph{Efficient long sequence modeling.}
The standard transformer architecture employs attention to comprehend the interplay between individual tokens. However, it faces challenges when dealing with lengthy sequences, primarily due to the quadratic computational complexity it entails. To address this issue, various attention approximation methods have been proposed \cite{hua2022transformer,ma2021luna,shen2021efficient,tay2020sparse,wang2020linformer,fei2022attention}, which aim to approximate self-attention while utilizing sub-quadratic computational resources. Notably, Mega \cite{ma2022mega} combines exponential moving average with a simplified attention unit, surpassing the performance of the baseline transformer models.
What's more, researchers have also explored alternatives that are capable of effectively handling long sequences. One involves employing state space models-based architectures, as exemplified by \cite{gu2021efficiently,gu2020hippo,gupta2022diagonal}, which have demonstrated significant advancements over contemporary state-of-the-art methods in tasks such as LRA and audio benchmarking \cite{goel2022s}. Moreover, recent studies \cite{gu2020hippo,peng2023rwkv,poli2023hyena,qin2024hierarchically} have provided empirical evidence supporting the potential of non-attention architectures in achieving commendable performance in language modeling. Motivated by this evolving trend of recurrence designs, our work draws inspiration from these advancements and predominantly leverages the backbone of RWKV.

\section{Conclusion}

This paper presents Diffusion-RWKV, an architecture designed for diffusion models featuring sequential information with linear computational complexity. The proposed approach effectively handles long-range hidden states without necessitating representation compression. Through comprehensive image generation tasks, we showcase its potential as a viable alternative backbone to the Transformer. Experimentally, Diffusion-RWKV demonstrates comparable performance and scalability while exhibiting lower computational complexity and memory consumption. Leveraging its reduced complexity, Diffusion-RWKV outperforms the Transformer model in scenarios where the latter struggles to cope with high computational demands. We anticipate that it will serve as an efficient and cost-effective substitute for the Transformer, thereby highlighting the substantial capabilities of transformers with linear complexity in the realm of multimodal generation. 
{
    \small
    \bibliographystyle{ieeenat_fullname}
    \bibliography{main}

\begin{thebibliography}{90}
\providecommand{\natexlab}[1]{#1}
\providecommand{\url}[1]{\texttt{#1}}
\expandafter\ifx\csname urlstyle\endcsname\relax
  \providecommand{\doi}[1]{doi: #1}\else
  \providecommand{\doi}{doi: \begingroup \urlstyle{rm}\Url}\fi

\bibitem[Bao et~al.(2023)Bao, Nie, Xue, Cao, Li, Su, and Zhu]{bao2023all}
Fan Bao, Shen Nie, Kaiwen Xue, Yue Cao, Chongxuan Li, Hang Su, and Jun Zhu.
\newblock All are worth words: A vit backbone for diffusion models.
\newblock In \emph{Proceedings of the IEEE/CVF Conference on Computer Vision and Pattern Recognition}, pages 22669--22679, 2023.

\bibitem[Brock et~al.(2018)Brock, Donahue, and Simonyan]{brock2018large}
Andrew Brock, Jeff Donahue, and Karen Simonyan.
\newblock Large scale gan training for high fidelity natural image synthesis.
\newblock \emph{arXiv preprint arXiv:1809.11096}, 2018.

\bibitem[Brown et~al.(2020)Brown, Mann, Ryder, Subbiah, Kaplan, Dhariwal, Neelakantan, Shyam, Sastry, Askell, et~al.]{brown2020language}
Tom Brown, Benjamin Mann, Nick Ryder, Melanie Subbiah, Jared~D Kaplan, Prafulla Dhariwal, Arvind Neelakantan, Pranav Shyam, Girish Sastry, Amanda Askell, et~al.
\newblock Language models are few-shot learners.
\newblock \emph{Advances in neural information processing systems}, 33:\penalty0 1877--1901, 2020.

\bibitem[Chang et~al.(2022)Chang, Zhang, Jiang, Liu, and Freeman]{chang2022maskgit}
Huiwen Chang, Han Zhang, Lu Jiang, Ce Liu, and William~T Freeman.
\newblock Maskgit: Masked generative image transformer.
\newblock In \emph{Proceedings of the IEEE/CVF Conference on Computer Vision and Pattern Recognition}, pages 11315--11325, 2022.

\bibitem[Deng et~al.(2009)Deng, Dong, Socher, Li, Li, and Fei-Fei]{deng2009imagenet}
Jia Deng, Wei Dong, Richard Socher, Li-Jia Li, Kai Li, and Li Fei-Fei.
\newblock Imagenet: A large-scale hierarchical image database.
\newblock In \emph{2009 IEEE conference on computer vision and pattern recognition}, pages 248--255. Ieee, 2009.

\bibitem[Devlin et~al.(2018)Devlin, Chang, Lee, and Toutanova]{devlin2018bert}
Jacob Devlin, Ming-Wei Chang, Kenton Lee, and Kristina Toutanova.
\newblock Bert: Pre-training of deep bidirectional transformers for language understanding.
\newblock \emph{arXiv preprint arXiv:1810.04805}, 2018.

\bibitem[Dhariwal and Nichol(2021)]{dhariwal2021diffusion}
Prafulla Dhariwal and Alexander Nichol.
\newblock Diffusion models beat gans on image synthesis.
\newblock \emph{Advances in neural information processing systems}, 34:\penalty0 8780--8794, 2021.

\bibitem[Ding et~al.(2021)Ding, Yang, Hong, Zheng, Zhou, Yin, Lin, Zou, Shao, Yang, et~al.]{ding2021cogview}
Ming Ding, Zhuoyi Yang, Wenyi Hong, Wendi Zheng, Chang Zhou, Da Yin, Junyang Lin, Xu Zou, Zhou Shao, Hongxia Yang, et~al.
\newblock Cogview: Mastering text-to-image generation via transformers.
\newblock \emph{Advances in Neural Information Processing Systems}, 34:\penalty0 19822--19835, 2021.

\bibitem[Ding et~al.(2022)Ding, Zheng, Hong, and Tang]{ding2022cogview2}
Ming Ding, Wendi Zheng, Wenyi Hong, and Jie Tang.
\newblock Cogview2: Faster and better text-to-image generation via hierarchical transformers.
\newblock \emph{Advances in Neural Information Processing Systems}, 35:\penalty0 16890--16902, 2022.

\bibitem[Dosovitskiy et~al.(2020)Dosovitskiy, Beyer, Kolesnikov, Weissenborn, Zhai, Unterthiner, Dehghani, Minderer, Heigold, Gelly, et~al.]{dosovitskiy2020image}
Alexey Dosovitskiy, Lucas Beyer, Alexander Kolesnikov, Dirk Weissenborn, Xiaohua Zhai, Thomas Unterthiner, Mostafa Dehghani, Matthias Minderer, Georg Heigold, Sylvain Gelly, et~al.
\newblock An image is worth 16x16 words: Transformers for image recognition at scale.
\newblock \emph{arXiv preprint arXiv:2010.11929}, 2020.

\bibitem[Duan et~al.(2024)Duan, Wang, Chen, Zhu, Lu, Lu, Qiao, Li, Dai, and Wang]{duan2024vision}
Yuchen Duan, Weiyun Wang, Zhe Chen, Xizhou Zhu, Lewei Lu, Tong Lu, Yu Qiao, Hongsheng Li, Jifeng Dai, and Wenhai Wang.
\newblock Vision-rwkv: Efficient and scalable visual perception with rwkv-like architectures.
\newblock \emph{arXiv preprint arXiv:2403.02308}, 2024.

\bibitem[Fei(2021)]{fei2021partially}
Zhengcong Fei.
\newblock Partially non-autoregressive image captioning.
\newblock In \emph{Proceedings of the AAAI Conference on Artificial Intelligence}, pages 1309--1316, 2021.

\bibitem[Fei(2022)]{fei2022attention}
Zhengcong Fei.
\newblock Attention-aligned transformer for image captioning.
\newblock In \emph{proceedings of the AAAI Conference on Artificial Intelligence}, pages 607--615, 2022.

\bibitem[Fei et~al.(2022{\natexlab{a}})Fei, Fan, Zhu, and Huang]{fei2022progressive}
Zhengcong Fei, Mingyuan Fan, Li Zhu, and Junshi Huang.
\newblock Progressive text-to-image generation.
\newblock \emph{arXiv preprint arXiv:2210.02291}, 2022{\natexlab{a}}.

\bibitem[Fei et~al.(2022{\natexlab{b}})Fei, Yan, Wang, and Tian]{fei2022deecap}
Zhengcong Fei, Xu Yan, Shuhui Wang, and Qi Tian.
\newblock Deecap: Dynamic early exiting for efficient image captioning.
\newblock In \emph{Proceedings of the IEEE/CVF Conference on Computer Vision and Pattern Recognition}, pages 12216--12226, 2022{\natexlab{b}}.

\bibitem[Fei et~al.(2023{\natexlab{a}})Fei, Fan, and Huang]{fei2023gradient}
Zhengcong Fei, Mingyuan Fan, and Junshi Huang.
\newblock Gradient-free textual inversion.
\newblock In \emph{Proceedings of the 31st ACM International Conference on Multimedia}, pages 1364--1373, 2023{\natexlab{a}}.

\bibitem[Fei et~al.(2023{\natexlab{b}})Fei, Fan, Zhu, Huang, Wei, and Wei]{fei2023uncertainty}
Zhengcong Fei, Mingyuan Fan, Li Zhu, Junshi Huang, Xiaoming Wei, and Xiaolin Wei.
\newblock Uncertainty-aware image captioning.
\newblock In \emph{Proceedings of the AAAI Conference on Artificial Intelligence}, pages 614--622, 2023{\natexlab{b}}.

\bibitem[Fei et~al.(2024)Fei, Fan, Yu, and Huang]{fei2024scalable}
Zhengcong Fei, Mingyuan Fan, Changqian Yu, and Junshi Huang.
\newblock Scalable diffusion models with state space backbone.
\newblock \emph{arXiv preprint arXiv:2402.05608}, 2024.

\bibitem[Fei(2019)]{fei2019fast}
Zheng-cong Fei.
\newblock Fast image caption generation with position alignment.
\newblock \emph{arXiv preprint arXiv:1912.06365}, 2019.

\bibitem[Goel et~al.(2022)Goel, Gu, Donahue, and R{\'e}]{goel2022s}
Karan Goel, Albert Gu, Chris Donahue, and Christopher R{\'e}.
\newblock It’s raw! audio generation with state-space models.
\newblock In \emph{International Conference on Machine Learning}, pages 7616--7633. PMLR, 2022.

\bibitem[Gu and Dao(2023)]{gu2023mamba}
Albert Gu and Tri Dao.
\newblock Mamba: Linear-time sequence modeling with selective state spaces.
\newblock \emph{arXiv preprint arXiv:2312.00752}, 2023.

\bibitem[Gu et~al.(2020)Gu, Dao, Ermon, Rudra, and R{\'e}]{gu2020hippo}
Albert Gu, Tri Dao, Stefano Ermon, Atri Rudra, and Christopher R{\'e}.
\newblock Hippo: Recurrent memory with optimal polynomial projections.
\newblock \emph{Advances in neural information processing systems}, 33:\penalty0 1474--1487, 2020.

\bibitem[Gu et~al.(2021)Gu, Goel, and R{\'e}]{gu2021efficiently}
Albert Gu, Karan Goel, and Christopher R{\'e}.
\newblock Efficiently modeling long sequences with structured state spaces.
\newblock \emph{arXiv preprint arXiv:2111.00396}, 2021.

\bibitem[Gupta et~al.(2022)Gupta, Gu, and Berant]{gupta2022diagonal}
Ankit Gupta, Albert Gu, and Jonathan Berant.
\newblock Diagonal state spaces are as effective as structured state spaces.
\newblock \emph{Advances in Neural Information Processing Systems}, 35:\penalty0 22982--22994, 2022.

\bibitem[He et~al.(2022)He, Chen, Xie, Li, Doll{\'a}r, and Girshick]{he2022masked}
Kaiming He, Xinlei Chen, Saining Xie, Yanghao Li, Piotr Doll{\'a}r, and Ross Girshick.
\newblock Masked autoencoders are scalable vision learners.
\newblock In \emph{Proceedings of the IEEE/CVF conference on computer vision and pattern recognition}, pages 16000--16009, 2022.

\bibitem[Heusel et~al.(2017)Heusel, Ramsauer, Unterthiner, Nessler, and Hochreiter]{heusel2017gans}
Martin Heusel, Hubert Ramsauer, Thomas Unterthiner, Bernhard Nessler, and Sepp Hochreiter.
\newblock Gans trained by a two time-scale update rule converge to a local nash equilibrium.
\newblock \emph{Advances in neural information processing systems}, 30, 2017.

\bibitem[Ho and Salimans(2022)]{ho2022classifier}
Jonathan Ho and Tim Salimans.
\newblock Classifier-free diffusion guidance.
\newblock \emph{arXiv preprint arXiv:2207.12598}, 2022.

\bibitem[Ho et~al.(2020)Ho, Jain, and Abbeel]{ho2020denoising}
Jonathan Ho, Ajay Jain, and Pieter Abbeel.
\newblock Denoising diffusion probabilistic models.
\newblock \emph{Advances in neural information processing systems}, 33:\penalty0 6840--6851, 2020.

\bibitem[Ho et~al.(2022)Ho, Saharia, Chan, Fleet, Norouzi, and Salimans]{ho2022cascaded}
Jonathan Ho, Chitwan Saharia, William Chan, David~J Fleet, Mohammad Norouzi, and Tim Salimans.
\newblock Cascaded diffusion models for high fidelity image generation.
\newblock \emph{The Journal of Machine Learning Research}, 23\penalty0 (1):\penalty0 2249--2281, 2022.

\bibitem[Hochreiter and Schmidhuber(1997)]{hochreiter1997long}
Sepp Hochreiter and J{\"u}rgen Schmidhuber.
\newblock Long short-term memory.
\newblock \emph{Neural computation}, 9\penalty0 (8):\penalty0 1735--1780, 1997.

\bibitem[Hu et~al.(2024)Hu, Baumann, Gui, Grebenkova, Ma, Fischer, and Ommer]{hu2024zigma}
Vincent~Tao Hu, Stefan~Andreas Baumann, Ming Gui, Olga Grebenkova, Pingchuan Ma, Johannes Fischer, and Bjorn Ommer.
\newblock Zigma: Zigzag mamba diffusion model.
\newblock \emph{arXiv preprint arXiv:2403.13802}, 2024.

\bibitem[Hua et~al.(2022)Hua, Dai, Liu, and Le]{hua2022transformer}
Weizhe Hua, Zihang Dai, Hanxiao Liu, and Quoc Le.
\newblock Transformer quality in linear time.
\newblock In \emph{International conference on machine learning}, pages 9099--9117. PMLR, 2022.

\bibitem[Hyv{\"a}rinen and Dayan(2005)]{hyvarinen2005estimation}
Aapo Hyv{\"a}rinen and Peter Dayan.
\newblock Estimation of non-normalized statistical models by score matching.
\newblock \emph{Journal of Machine Learning Research}, 6\penalty0 (4), 2005.

\bibitem[Karras et~al.(2022)Karras, Aittala, Aila, and Laine]{karras2022elucidating}
Tero Karras, Miika Aittala, Timo Aila, and Samuli Laine.
\newblock Elucidating the design space of diffusion-based generative models.
\newblock \emph{Advances in Neural Information Processing Systems}, 35:\penalty0 26565--26577, 2022.

\bibitem[Kim et~al.(2021)Kim, Shin, Song, Kang, and Moon]{kim2021soft}
Dongjun Kim, Seungjae Shin, Kyungwoo Song, Wanmo Kang, and Il-Chul Moon.
\newblock Soft truncation: A universal training technique of score-based diffusion model for high precision score estimation.
\newblock \emph{arXiv preprint arXiv:2106.05527}, 2021.

\bibitem[Kingma and Ba(2014)]{kingma2014adam}
Diederik~P Kingma and Jimmy Ba.
\newblock Adam: A method for stochastic optimization.
\newblock \emph{arXiv preprint arXiv:1412.6980}, 2014.

\bibitem[Kingma and Gao(2023)]{kingma2023understanding}
Diederik~P Kingma and Ruiqi Gao.
\newblock Understanding diffusion objectives as the elbo with simple data augmentation.
\newblock In \emph{Thirty-seventh Conference on Neural Information Processing Systems}, 2023.

\bibitem[Kingma and Welling(2013)]{kingma2013auto}
Diederik~P Kingma and Max Welling.
\newblock Auto-encoding variational bayes.
\newblock \emph{arXiv preprint arXiv:1312.6114}, 2013.

\bibitem[Kitaev et~al.(2020)Kitaev, Kaiser, and Levskaya]{kitaev2020reformer}
Nikita Kitaev, {\L}ukasz Kaiser, and Anselm Levskaya.
\newblock Reformer: The efficient transformer.
\newblock \emph{arXiv preprint arXiv:2001.04451}, 2020.

\bibitem[Krizhevsky et~al.(2009)Krizhevsky, Hinton, et~al.]{krizhevsky2009learning}
Alex Krizhevsky, Geoffrey Hinton, et~al.
\newblock Learning multiple layers of features from tiny images.
\newblock 2009.

\bibitem[Kynk{\"a}{\"a}nniemi et~al.(2019)Kynk{\"a}{\"a}nniemi, Karras, Laine, Lehtinen, and Aila]{kynkaanniemi2019improved}
Tuomas Kynk{\"a}{\"a}nniemi, Tero Karras, Samuli Laine, Jaakko Lehtinen, and Timo Aila.
\newblock Improved precision and recall metric for assessing generative models.
\newblock \emph{Advances in Neural Information Processing Systems}, 32, 2019.

\bibitem[Lee et~al.(2022)Lee, Kim, Kim, Cho, and Han]{lee2022autoregressive}
Doyup Lee, Chiheon Kim, Saehoon Kim, Minsu Cho, and Wook-Shin Han.
\newblock Autoregressive image generation using residual quantization.
\newblock In \emph{Proceedings of the IEEE/CVF Conference on Computer Vision and Pattern Recognition}, pages 11523--11532, 2022.

\bibitem[Lewis et~al.(2019)Lewis, Liu, Goyal, Ghazvininejad, Mohamed, Levy, Stoyanov, and Zettlemoyer]{lewis2019bart}
Mike Lewis, Yinhan Liu, Naman Goyal, Marjan Ghazvininejad, Abdelrahman Mohamed, Omer Levy, Ves Stoyanov, and Luke Zettlemoyer.
\newblock Bart: Denoising sequence-to-sequence pre-training for natural language generation, translation, and comprehension.
\newblock \emph{arXiv preprint arXiv:1910.13461}, 2019.

\bibitem[Lin et~al.(2022)Lin, Wang, Liu, and Qiu]{lin2022survey}
Tianyang Lin, Yuxin Wang, Xiangyang Liu, and Xipeng Qiu.
\newblock A survey of transformers.
\newblock \emph{AI open}, 3:\penalty0 111--132, 2022.

\bibitem[Liu et~al.(2019)Liu, Ott, Goyal, Du, Joshi, Chen, Levy, Lewis, Zettlemoyer, and Stoyanov]{liu2019roberta}
Yinhan Liu, Myle Ott, Naman Goyal, Jingfei Du, Mandar Joshi, Danqi Chen, Omer Levy, Mike Lewis, Luke Zettlemoyer, and Veselin Stoyanov.
\newblock Roberta: A robustly optimized bert pretraining approach.
\newblock \emph{arXiv preprint arXiv:1907.11692}, 2019.

\bibitem[Liu et~al.(2015)Liu, Luo, Wang, and Tang]{liu2015deep}
Ziwei Liu, Ping Luo, Xiaogang Wang, and Xiaoou Tang.
\newblock Deep learning face attributes in the wild.
\newblock In \emph{Proceedings of the IEEE international conference on computer vision}, pages 3730--3738, 2015.

\bibitem[Liu et~al.(2021)Liu, Lin, Cao, Hu, Wei, Zhang, Lin, and Guo]{liu2021swin}
Ze Liu, Yutong Lin, Yue Cao, Han Hu, Yixuan Wei, Zheng Zhang, Stephen Lin, and Baining Guo.
\newblock Swin transformer: Hierarchical vision transformer using shifted windows.
\newblock In \emph{Proceedings of the IEEE/CVF international conference on computer vision}, pages 10012--10022, 2021.

\bibitem[Lu et~al.(2022{\natexlab{a}})Lu, Zhou, Bao, Chen, Li, and Zhu]{lu2022dpm}
Cheng Lu, Yuhao Zhou, Fan Bao, Jianfei Chen, Chongxuan Li, and Jun Zhu.
\newblock Dpm-solver: A fast ode solver for diffusion probabilistic model sampling in around 10 steps.
\newblock \emph{Advances in Neural Information Processing Systems}, 35:\penalty0 5775--5787, 2022{\natexlab{a}}.

\bibitem[Lu et~al.(2022{\natexlab{b}})Lu, Zhou, Bao, Chen, Li, and Zhu]{lu2022dpm+}
Cheng Lu, Yuhao Zhou, Fan Bao, Jianfei Chen, Chongxuan Li, and Jun Zhu.
\newblock Dpm-solver++: Fast solver for guided sampling of diffusion probabilistic models.
\newblock \emph{arXiv preprint arXiv:2211.01095}, 2022{\natexlab{b}}.

\bibitem[Ma et~al.(2024)Ma, Goldstein, Albergo, Boffi, Vanden-Eijnden, and Xie]{ma2024sit}
Nanye Ma, Mark Goldstein, Michael~S Albergo, Nicholas~M Boffi, Eric Vanden-Eijnden, and Saining Xie.
\newblock Sit: Exploring flow and diffusion-based generative models with scalable interpolant transformers.
\newblock \emph{arXiv preprint arXiv:2401.08740}, 2024.

\bibitem[Ma et~al.(2021)Ma, Kong, Wang, Zhou, May, Ma, and Zettlemoyer]{ma2021luna}
Xuezhe Ma, Xiang Kong, Sinong Wang, Chunting Zhou, Jonathan May, Hao Ma, and Luke Zettlemoyer.
\newblock Luna: Linear unified nested attention.
\newblock \emph{Advances in Neural Information Processing Systems}, 34:\penalty0 2441--2453, 2021.

\bibitem[Ma et~al.(2022)Ma, Zhou, Kong, He, Gui, Neubig, May, and Zettlemoyer]{ma2022mega}
Xuezhe Ma, Chunting Zhou, Xiang Kong, Junxian He, Liangke Gui, Graham Neubig, Jonathan May, and Luke Zettlemoyer.
\newblock Mega: moving average equipped gated attention.
\newblock \emph{arXiv preprint arXiv:2209.10655}, 2022.

\bibitem[Mann et~al.(2020)Mann, Ryder, Subbiah, Kaplan, Dhariwal, Neelakantan, Shyam, Sastry, Askell, Agarwal, et~al.]{mann2020language}
Ben Mann, N Ryder, M Subbiah, J Kaplan, P Dhariwal, A Neelakantan, P Shyam, G Sastry, A Askell, S Agarwal, et~al.
\newblock Language models are few-shot learners.
\newblock \emph{arXiv preprint arXiv:2005.14165}, 2020.

\bibitem[Nash et~al.(2021)Nash, Menick, Dieleman, and Battaglia]{nash2021generating}
Charlie Nash, Jacob Menick, Sander Dieleman, and Peter~W Battaglia.
\newblock Generating images with sparse representations.
\newblock \emph{arXiv preprint arXiv:2103.03841}, 2021.

\bibitem[Nichol and Dhariwal(2021)]{nichol2021improved}
Alexander~Quinn Nichol and Prafulla Dhariwal.
\newblock Improved denoising diffusion probabilistic models.
\newblock In \emph{International Conference on Machine Learning}, pages 8162--8171. PMLR, 2021.

\bibitem[Parmar et~al.(2022)Parmar, Zhang, and Zhu]{parmar2022aliased}
Gaurav Parmar, Richard Zhang, and Jun-Yan Zhu.
\newblock On aliased resizing and surprising subtleties in gan evaluation.
\newblock In \emph{Proceedings of the IEEE/CVF Conference on Computer Vision and Pattern Recognition}, pages 11410--11420, 2022.

\bibitem[Parmar et~al.(2018)Parmar, Vaswani, Uszkoreit, Kaiser, Shazeer, Ku, and Tran]{parmar2018image}
Niki Parmar, Ashish Vaswani, Jakob Uszkoreit, Lukasz Kaiser, Noam Shazeer, Alexander Ku, and Dustin Tran.
\newblock Image transformer.
\newblock In \emph{International conference on machine learning}, pages 4055--4064. PMLR, 2018.

\bibitem[Peebles and Xie(2023)]{peebles2023scalable}
William Peebles and Saining Xie.
\newblock Scalable diffusion models with transformers.
\newblock In \emph{Proceedings of the IEEE/CVF International Conference on Computer Vision}, pages 4195--4205, 2023.

\bibitem[Peng et~al.(2023)Peng, Alcaide, Anthony, Albalak, Arcadinho, Cao, Cheng, Chung, Grella, GV, et~al.]{peng2023rwkv}
Bo Peng, Eric Alcaide, Quentin Anthony, Alon Albalak, Samuel Arcadinho, Huanqi Cao, Xin Cheng, Michael Chung, Matteo Grella, Kranthi~Kiran GV, et~al.
\newblock Rwkv: Reinventing rnns for the transformer era.
\newblock \emph{arXiv preprint arXiv:2305.13048}, 2023.

\bibitem[Perez et~al.(2018)Perez, Strub, De~Vries, Dumoulin, and Courville]{perez2018film}
Ethan Perez, Florian Strub, Harm De~Vries, Vincent Dumoulin, and Aaron Courville.
\newblock Film: Visual reasoning with a general conditioning layer.
\newblock In \emph{Proceedings of the AAAI conference on artificial intelligence}, 2018.

\bibitem[Poli et~al.(2023)Poli, Massaroli, Nguyen, Fu, Dao, Baccus, Bengio, Ermon, and R{\'e}]{poli2023hyena}
Michael Poli, Stefano Massaroli, Eric Nguyen, Daniel~Y Fu, Tri Dao, Stephen Baccus, Yoshua Bengio, Stefano Ermon, and Christopher R{\'e}.
\newblock Hyena hierarchy: Towards larger convolutional language models.
\newblock In \emph{International Conference on Machine Learning}, pages 28043--28078. PMLR, 2023.

\bibitem[Qin et~al.(2024)Qin, Yang, and Zhong]{qin2024hierarchically}
Zhen Qin, Songlin Yang, and Yiran Zhong.
\newblock Hierarchically gated recurrent neural network for sequence modeling.
\newblock \emph{Advances in Neural Information Processing Systems}, 36, 2024.

\bibitem[Radford et~al.(2018)Radford, Narasimhan, Salimans, Sutskever, et~al.]{radford2018improving}
Alec Radford, Karthik Narasimhan, Tim Salimans, Ilya Sutskever, et~al.
\newblock Improving language understanding by generative pre-training.
\newblock 2018.

\bibitem[Radford et~al.(2019)Radford, Wu, Child, Luan, Amodei, Sutskever, et~al.]{radford2019language}
Alec Radford, Jeffrey Wu, Rewon Child, David Luan, Dario Amodei, Ilya Sutskever, et~al.
\newblock Language models are unsupervised multitask learners.
\newblock \emph{OpenAI blog}, 1\penalty0 (8):\penalty0 9, 2019.

\bibitem[Ramesh et~al.(2021)Ramesh, Pavlov, Goh, Gray, Voss, Radford, Chen, and Sutskever]{ramesh2021zero}
Aditya Ramesh, Mikhail Pavlov, Gabriel Goh, Scott Gray, Chelsea Voss, Alec Radford, Mark Chen, and Ilya Sutskever.
\newblock Zero-shot text-to-image generation.
\newblock In \emph{International conference on machine learning}, pages 8821--8831. Pmlr, 2021.

\bibitem[Ramesh et~al.(2022)Ramesh, Dhariwal, Nichol, Chu, and Chen]{ramesh2022hierarchical}
Aditya Ramesh, Prafulla Dhariwal, Alex Nichol, Casey Chu, and Mark Chen.
\newblock Hierarchical text-conditional image generation with clip latents.
\newblock \emph{arXiv preprint arXiv:2204.06125}, 1\penalty0 (2):\penalty0 3, 2022.

\bibitem[Rombach et~al.(2022)Rombach, Blattmann, Lorenz, Esser, and Ommer]{rombach2022high}
Robin Rombach, Andreas Blattmann, Dominik Lorenz, Patrick Esser, and Bj{\"o}rn Ommer.
\newblock High-resolution image synthesis with latent diffusion models.
\newblock In \emph{Proceedings of the IEEE/CVF conference on computer vision and pattern recognition}, pages 10684--10695, 2022.

\bibitem[Saharia et~al.(2022)Saharia, Chan, Saxena, Li, Whang, Denton, Ghasemipour, Gontijo~Lopes, Karagol~Ayan, Salimans, et~al.]{saharia2022photorealistic}
Chitwan Saharia, William Chan, Saurabh Saxena, Lala Li, Jay Whang, Emily~L Denton, Kamyar Ghasemipour, Raphael Gontijo~Lopes, Burcu Karagol~Ayan, Tim Salimans, et~al.
\newblock Photorealistic text-to-image diffusion models with deep language understanding.
\newblock \emph{Advances in Neural Information Processing Systems}, 35:\penalty0 36479--36494, 2022.

\bibitem[Salimans et~al.(2016)Salimans, Goodfellow, Zaremba, Cheung, Radford, and Chen]{salimans2016improved}
Tim Salimans, Ian Goodfellow, Wojciech Zaremba, Vicki Cheung, Alec Radford, and Xi Chen.
\newblock Improved techniques for training gans.
\newblock \emph{Advances in neural information processing systems}, 29, 2016.

\bibitem[Sauer et~al.(2022)Sauer, Schwarz, and Geiger]{sauer2022stylegan}
Axel Sauer, Katja Schwarz, and Andreas Geiger.
\newblock Stylegan-xl: Scaling stylegan to large diverse datasets.
\newblock In \emph{ACM SIGGRAPH 2022 conference proceedings}, pages 1--10, 2022.

\bibitem[Shaham et~al.(2018)Shaham, Stanton, Li, Nadler, Basri, and Kluger]{shaham2018spectralnet}
Uri Shaham, Kelly Stanton, Henry Li, Boaz Nadler, Ronen Basri, and Yuval Kluger.
\newblock Spectralnet: Spectral clustering using deep neural networks.
\newblock \emph{arXiv preprint arXiv:1801.01587}, 2018.

\bibitem[Shen et~al.(2021)Shen, Zhang, Zhao, Yi, and Li]{shen2021efficient}
Zhuoran Shen, Mingyuan Zhang, Haiyu Zhao, Shuai Yi, and Hongsheng Li.
\newblock Efficient attention: Attention with linear complexities.
\newblock In \emph{Proceedings of the IEEE/CVF winter conference on applications of computer vision}, pages 3531--3539, 2021.

\bibitem[Smith et~al.(2022)Smith, Patwary, Norick, LeGresley, Rajbhandari, Casper, Liu, Prabhumoye, Zerveas, Korthikanti, et~al.]{smith2022using}
Shaden Smith, Mostofa Patwary, Brandon Norick, Patrick LeGresley, Samyam Rajbhandari, Jared Casper, Zhun Liu, Shrimai Prabhumoye, George Zerveas, Vijay Korthikanti, et~al.
\newblock Using deepspeed and megatron to train megatron-turing nlg 530b, a large-scale generative language model.
\newblock \emph{arXiv preprint arXiv:2201.11990}, 2022.

\bibitem[Song et~al.(2020{\natexlab{a}})Song, Meng, and Ermon]{song2020denoising}
Jiaming Song, Chenlin Meng, and Stefano Ermon.
\newblock Denoising diffusion implicit models.
\newblock \emph{arXiv preprint arXiv:2010.02502}, 2020{\natexlab{a}}.

\bibitem[Song and Ermon(2019)]{song2019generative}
Yang Song and Stefano Ermon.
\newblock Generative modeling by estimating gradients of the data distribution.
\newblock \emph{Advances in neural information processing systems}, 32, 2019.

\bibitem[Song et~al.(2020{\natexlab{b}})Song, Sohl-Dickstein, Kingma, Kumar, Ermon, and Poole]{song2020score}
Yang Song, Jascha Sohl-Dickstein, Diederik~P Kingma, Abhishek Kumar, Stefano Ermon, and Ben Poole.
\newblock Score-based generative modeling through stochastic differential equations.
\newblock \emph{arXiv preprint arXiv:2011.13456}, 2020{\natexlab{b}}.

\bibitem[Song et~al.(2023)Song, Dhariwal, Chen, and Sutskever]{song2023consistency}
Yang Song, Prafulla Dhariwal, Mark Chen, and Ilya Sutskever.
\newblock Consistency models.
\newblock \emph{arXiv preprint arXiv:2303.01469}, 2023.

\bibitem[Stickland and Murray(2019)]{stickland2019bert}
Asa~Cooper Stickland and Iain Murray.
\newblock Bert and pals: Projected attention layers for efficient adaptation in multi-task learning.
\newblock In \emph{International Conference on Machine Learning}, pages 5986--5995. PMLR, 2019.

\bibitem[Tay et~al.(2020)Tay, Bahri, Yang, Metzler, and Juan]{tay2020sparse}
Yi Tay, Dara Bahri, Liu Yang, Donald Metzler, and Da-Cheng Juan.
\newblock Sparse sinkhorn attention.
\newblock In \emph{International Conference on Machine Learning}, pages 9438--9447. PMLR, 2020.

\bibitem[Tay et~al.(2022)Tay, Dehghani, Bahri, and Metzler]{tay2022efficient}
Yi Tay, Mostafa Dehghani, Dara Bahri, and Donald Metzler.
\newblock Efficient transformers: A survey.
\newblock \emph{ACM Computing Surveys}, 55\penalty0 (6):\penalty0 1--28, 2022.

\bibitem[Vaswani et~al.(2017)Vaswani, Shazeer, Parmar, Uszkoreit, Jones, Gomez, Kaiser, and Polosukhin]{vaswani2017attention}
Ashish Vaswani, Noam Shazeer, Niki Parmar, Jakob Uszkoreit, Llion Jones, Aidan~N Gomez, {\L}ukasz Kaiser, and Illia Polosukhin.
\newblock Attention is all you need.
\newblock \emph{Advances in neural information processing systems}, 30, 2017.

\bibitem[Wang et~al.(2020)Wang, Li, Khabsa, Fang, and Ma]{wang2020linformer}
Sinong Wang, Belinda~Z Li, Madian Khabsa, Han Fang, and Hao Ma.
\newblock Linformer: Self-attention with linear complexity.
\newblock \emph{arXiv preprint arXiv:2006.04768}, 2020.

\bibitem[Wu and He(2018)]{wu2018group}
Yuxin Wu and Kaiming He.
\newblock Group normalization.
\newblock In \emph{Proceedings of the European conference on computer vision (ECCV)}, pages 3--19, 2018.

\bibitem[Yan et~al.(2023)Yan, Gu, and Rush]{yan2023diffusion}
Jing~Nathan Yan, Jiatao Gu, and Alexander~M Rush.
\newblock Diffusion models without attention.
\newblock \emph{arXiv preprint arXiv:2311.18257}, 2023.

\bibitem[Yan et~al.(2021)Yan, Fei, Li, Wang, Huang, and Tian]{yan2021semi}
Xu Yan, Zhengcong Fei, Zekang Li, Shuhui Wang, Qingming Huang, and Qi Tian.
\newblock Semi-autoregressive image captioning.
\newblock In \emph{Proceedings of the 29th ACM International Conference on Multimedia}, pages 2708--2716, 2021.

\bibitem[Yang et~al.(2022)Yang, Shih, Fu, Zhao, and Ji]{yang2022your}
Xiulong Yang, Sheng-Min Shih, Yinlin Fu, Xiaoting Zhao, and Shihao Ji.
\newblock Your vit is secretly a hybrid discriminative-generative diffusion model.
\newblock \emph{arXiv preprint arXiv:2208.07791}, 2022.

\bibitem[Zamir et~al.(2022)Zamir, Arora, Khan, Hayat, Khan, and Yang]{zamir2022restormer}
Syed~Waqas Zamir, Aditya Arora, Salman Khan, Munawar Hayat, Fahad~Shahbaz Khan, and Ming-Hsuan Yang.
\newblock Restormer: Efficient transformer for high-resolution image restoration.
\newblock In \emph{Proceedings of the IEEE/CVF conference on computer vision and pattern recognition}, pages 5728--5739, 2022.

\bibitem[Zheng et~al.(2023)Zheng, Nie, Vahdat, and Anandkumar]{zheng2023fast}
Hongkai Zheng, Weili Nie, Arash Vahdat, and Anima Anandkumar.
\newblock Fast training of diffusion models with masked transformers.
\newblock \emph{arXiv preprint arXiv:2306.09305}, 2023.

\bibitem[Zhou et~al.(2021)Zhou, Zhang, Peng, Zhang, Li, Xiong, and Zhang]{zhou2021informer}
Haoyi Zhou, Shanghang Zhang, Jieqi Peng, Shuai Zhang, Jianxin Li, Hui Xiong, and Wancai Zhang.
\newblock Informer: Beyond efficient transformer for long sequence time-series forecasting.
\newblock In \emph{Proceedings of the AAAI conference on artificial intelligence}, pages 11106--11115, 2021.

\bibitem[Zhu et~al.(2024)Zhu, Liao, Zhang, Wang, Liu, and Wang]{zhu2024vision}
Lianghui Zhu, Bencheng Liao, Qian Zhang, Xinlong Wang, Wenyu Liu, and Xinggang Wang.
\newblock Vision mamba: Efficient visual representation learning with bidirectional state space model.
\newblock \emph{arXiv preprint arXiv:2401.09417}, 2024.

\end{thebibliography}
}


\end{document}